%% file: main.tex
\documentclass{article}

\usepackage{microtype}
\usepackage{graphicx}
\usepackage{subfigure}
\usepackage{booktabs} 
\input{defs}

\usepackage{hyperref}

\newcommand{\AIResidency}{\textsuperscript{*}Work done as part of the Google AI Residency Program.}

\usepackage[accepted]{icml2019}

\def\title{Learning to Generalize from Sparse and Underspecified Rewards}

\icmltitlerunning{\title}

\begin{document}

\twocolumn[
\icmltitle{\title}



\icmlsetsymbol{aires}{*}

\begin{icmlauthorlist}
\icmlauthor{Rishabh Agarwal}{goo,aires}
\icmlauthor{Chen Liang}{goo}
\icmlauthor{Dale Schuurmans}{goo,ab}
\icmlauthor{Mohammad Norouzi}{goo}
\end{icmlauthorlist}

\icmlaffiliation{goo}{Google Research, Brain Team}
\icmlaffiliation{ab}{University of Alberta}

\icmlcorrespondingauthor{Rishabh Agarwal}{rishabhagarwal@google.com}
\icmlcorrespondingauthor{Mohammad Norouzi}{mnorouzi@google.com}

\icmlkeywords{NLP, Meta Learning, Semantic Parsing, Reinforcement Learning}

\vskip 0.3in
]



\printAffiliationsAndNotice{\AIResidency} 

\begin{abstract}
\input{abs}
\end{abstract}

\section{Introduction}
\input{intro}

\section{Formulation}
\input{background}

\section{Mode Covering Exploration (MAPOX)}
\input{method1}

\section{Learning Rewards without Demonstration}
\input{method2}

\section{Related Work}
\input{related}

\section{Experiments}
\input{experiments}
\input{results}


\section{Conclusion \& Future Work}
\input{conc}

\input{acknowledgment.tex}


\bibliography{main}
\bibliographystyle{icml2019}

\appendix
\cleardoublepage
\icmltitlerunning{Supplementary Material for \title}
\input{sup}

\end{document}

%% file: defs.tex
\usepackage{dsfont}
\usepackage{amsthm}
\usepackage{amsmath}
\usepackage{amssymb}
\usepackage{color}
\usepackage{xspace}
\usepackage{pgfplots, pgfplotstable}
\usepackage{xcolor,colortbl}

\newcommand{\argmax}[1]{\underset{#1}{\operatorname{argmax}}}

\renewcommand{\mid}[0]{\:\vert\:}

\def\triangle{\raisebox{0.1ex}{$\blacktriangleright$}~}

\def\vphi{V_\phi}

\newcommand{\R}[1]{R}
\newcommand{\G}[1]{G}

\newcommand{\Ad}[1]{A}

\newcommand{\CC}[1]{C}

\def\setA{\mathcal{A}}
\def\setAp{\mathcal{A}^+}

\def\setBp{\mathcal{B}^+}
\def\setBpT{\mathcal{B}^{+}_{\mathrm{train}}}
\def\setBpV{\mathcal{B}^{+}_{\mathrm{val}}}

\def\setDT{\mathcal{D}_{\mathrm{train}}}
\def\setDV{\mathcal{D}_{\mathrm{val}}}
\def\setV{\mathcal{V}}
\def\setS{\mathcal{S}}
\def\setX{\mathcal{X}}
\def\vthetaP{\vtheta^{\prime}}

\def\eg{{\em e.g.,}}
\def\ie{{\em i.e.,}}

\newcommand{\figref}[1]{Figure~\ref{#1}}

\newcommand{\algref}[1]{Algorithm~\ref{#1}}
\newcommand{\tabref}[1]{Table~\ref{#1}}

\newcommand{\one}[1]{\mathds{1}[#1]}

\newcommand{\kl}[2]{{\mathrm{KL}}\left(#1~\Vert~#2\right)}

\usepackage{amsmath,amsfonts,bm}

\def\1{\bm{1}}

\def\vtheta{{\bm{\theta}}}
\def\vphi{{\bm{\phi}}}
\def\va{{\bm{a}}}

\def\vc{{\bm{c}}}

\def\vf{{\bm{f}}}

\def\vj{{\bm{j}}}

\def\vw{{\bm{w}}}
\def\vx{{\bm{x}}}
\def\vy{{\bm{a}}}
\def\vyp{{\bm{a}^+}}

\DeclareMathAlphabet{\mathsfit}{\encodingdefault}{\sfdefault}{m}{sl}
\SetMathAlphabet{\mathsfit}{bold}{\encodingdefault}{\sfdefault}{bx}{n}

\def\sD{{\mathcal{D}}}

\def\ww{\hspace*{.075cm}}
\def\www{\hspace*{.4cm}}
\newcommand{\script}[1]{\scriptsize {(#1)}}

%% file: abs.tex
We consider the problem of learning from sparse and underspecified
rewards, where an agent receives a complex input, such as a natural
language instruction, and needs to generate a complex response, such
as an action sequence, while only receiving binary success-failure
feedback.  Such success-failure rewards are often underspecified: they
do not distinguish between purposeful and accidental
success. Generalization from underspecified rewards hinges on
discounting spurious trajectories that attain accidental success,
while learning from sparse feedback requires effective exploration.
We address exploration by using a mode covering direction of KL
divergence to collect a diverse set of successful trajectories,
followed by a mode seeking KL divergence to train a robust policy.  We
propose Meta Reward Learning (MeRL) to construct an auxiliary reward
function that provides more refined feedback for learning.  The
parameters of the auxiliary reward function are optimized with respect
to the validation performance of a trained policy. The MeRL approach
outperforms an alternative method for reward learning based on
Bayesian Optimization, and achieves the state-of-the-art on
weakly-supervised semantic parsing. It improves previous work by 1.2\%
and 2.4\% on \textsc{WikiTableQuestions} and \textsc{WikiSQL} datasets
respectively.

%% file: intro.tex
Effortlessly communicating with computers using natural language has
been a longstanding goal of artificial intelligence~\cite{Winograd71}.
Reinforcement Learning (RL) presents a flexible framework for
optimizing goal oriented behavior~\cite{sutton18}. As such, one can
use RL to optimize language communication if it is expressed in terms
of achieving concrete goals.  In this pursuit, researchers have
created a number of simulation environments where a learning agent is
provided with a natural language input and asked to produce a sequence
of actions for achieving a goal specified in the input
text~(\eg~\citet{long16,hermann17,chaplot18,fu19,babyai}).  These
tasks are typically episodic, where the agent receives sparse binary
success-failure feedback indicating whether an intended goal has been
accomplished.  After training, the agent is placed in new contexts and
evaluated based on its ability to reach novel goals, indicating the quality
of its behavior policy and language interpretation skills. The
emphasis on generalization in these tasks makes them
suitable for benchmarking overfitting in
RL~\cite{cobbe2018quantifying, zhang2018study}.

\begin{figure}[t]
\input{fig1}
\caption{{\em Semantic parsing} from question-answer pairs. An agent is presented with a natural language question $\vx$
and is asked to generate a SQL-like program $\va$. The agent receives
a reward of $1$ if execution of a program $\va$ on the relevant data
table leads to the correct answer (\eg~USA).  The reward is
underspecified because {\em spurious} programs (\eg~$\va_2, \va_3$) can
also achieve a reward of $1$. }
\label{fig:fig1}
\end{figure}
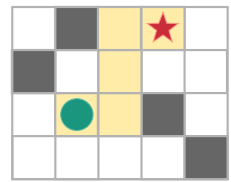

\begin{figure}[t]
\input{fig2}
\caption{{\em Instruction following} in a simple maze. A blind agent is presented with
a sequence of (Left, Right, Up, Down) instructions. Given the input
text, the agent ($\bullet$) performs a sequence of actions, and only
receives a reward of $1$ if it reaches the goal ($\star$).}
\label{fig:fig2}
\vspace*{-0.15in}
\end{figure}

\figref{fig:fig1} and \ref{fig:fig2} illustrate two examples
of contextual environments with sparse and underspecified rewards.
The rewards are {\em sparse}, since only a few trajectories in the
combinatorial space of all trajectories leads to a non-zero return.
In addition, the rewards are {\em underspecified}, since the agent may
receive a return of $1$ for exploiting {\em spurious} patterns in the
environment.  We assert that the generalization performance of an
agent trained in this setting hinges on (1) effective exploration to
find successful trajectories, and (2) discounting spurious
trajectories to learn a generalizable behavior.

To facilitate effective and principled exploration,
we propose to disentangle combinatorial search and exploration
from robust policy optimization. In particular, we use a {\em mode
covering} direction of KL divergence to learn a high entropy
exploration policy to help collect a diverse set of successful
trajectories.
Then, given a buffer of promising trajectories,
we use a {\em mode seeking} direction of KL divergence
to learn a robust policy
with favorable generalization performance.

A key challenge in language conditional learning environments is the
lack of fully specified rewards that perfectly distinguish optimal
and suboptimal trajectories.  Designing a rich trajectory-level
reward function requires a deep understanding of the semantic
relationship between the environment and the natural language input,
which is not available in most real-world settings.  Such a challenge
arises in weakly supervised semantic parsing as depicted
in \figref{fig:fig1}~\cite{pasupat2015tables}.  From an AI safety
perspective, underspecified rewards may lead to reward
hacking~\cite{amodei2016concrete} causing unintended and harmful
behavior when deployed in real-world scenarios.

In this paper, we investigate whether one can automatically discover a
rich trajectory-level reward function to help a learning agent
discount spurious trajectories and improve generalization.  Toward
this end, we utilize both gradient-based
Meta-Learning~\cite{finn2017model,maclaurin2015gradient} and Bayesian
Optimization~\cite{snoek2012practical} for reward learning. We propose
to optimize the parameters of the auxiliary reward function in an
outer loop to maximize generalization performance of a policy trained
based on the auxiliary rewards. Our work is distinct from recent
works~\cite{bahdanau19, fu19} on learning rewards for language tasks because
we do not require any form of trajectory or goal demonstration.

\begin{figure}[t]
\begin{center}
\includegraphics[width=1.0\columnwidth]{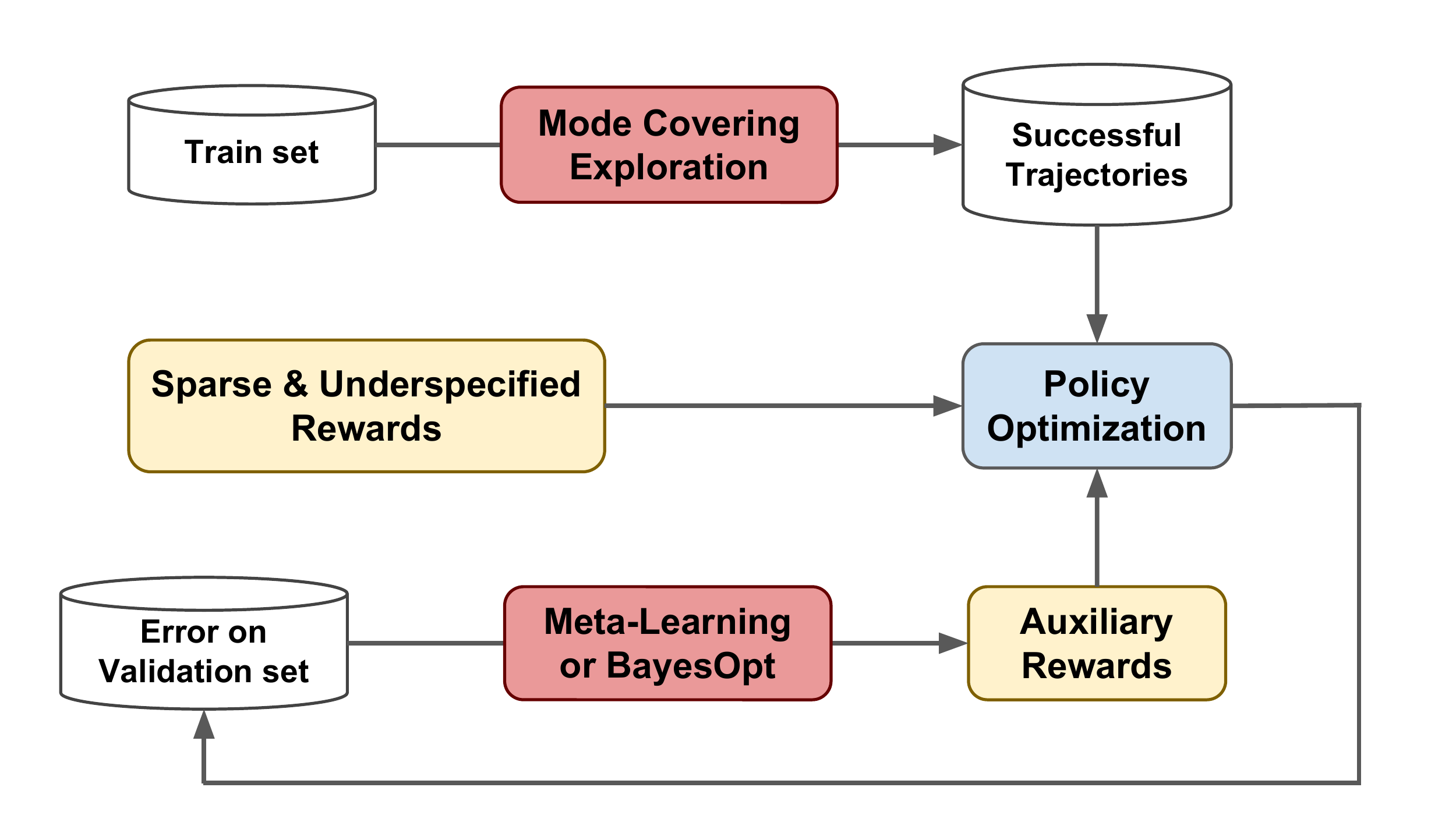}
\end{center}
\vspace{-.15in}
\caption{Overview of the proposed approach. We employ (1) mode covering exploration to collect a diverse set of successful trajectories in a memory buffer;
(2) Meta-learning or Bayesian optimization to learn an auxiliary reward function to discount spurious trajectories.}
\label{fig:fig3}
\vspace{-.05in}
\end{figure}

We evaluate our overall approach (see Figure \ref{fig:fig3} for an
overview) on two real world weakly-supervised semantic parsing
benchmarks~\cite{pasupat2015tables,zhong2017seq2sql}
(\figref{fig:fig1}) and a simple instruction following environment
(\figref{fig:fig2}).  In all of the experiments, we observe a
significant benefit from the proposed Meta Reward Learning~(MeRL)
approach, even when the exploration problem is synthetically
mitigated.  In addition, we achieve notable gains from the mode
covering exploration strategy, which combines well with MeRL to
achieve the state-of-the-art results on weakly-supervised semantic
parsing.

%% file: fig1.tex
\small
\begin{tabular}{@{}l@{\hspace*{.3cm}}l@{}}
\begin{tabular}{@{}|@{\ww}l@{\ww}l@{\ww}l@{\ww}l@{\ww}|@{}}
\hline
Rank & Nation & Gold & Silver \\
\hline
1 & USA	&	10 &	12 \\
2 & GBR	&	9 &	4 \\
3 & CHN	&	8 &	11 \\
4 & RUS &	2 &	4 \\
5 & GER &	2 &	2 \\
6 & JPN & 	2 &	1 \\
7 & FRA &	2 &	1 \\
\hline
\end{tabular}%
&
\raisebox{1.2cm}{\begin{minipage}[t]{4.65cm}
$\vx$ = ``Which nation won the most\\silver medal?''\\
$R(\va) = \one{\texttt{Execute}(\va) = \text{``USA''}}$\\[.3cm]
$\va_1$ = \texttt{argmax\_row(}Silver\texttt{)}.Nation\\
$\va_2$ = \texttt{argmax\_row(}Gold\texttt{)}.Nation\\
$\va_3$ = \texttt{argmin\_row(}Rank\texttt{)}.Nation\\
$R(\va_1) = R(\va_2) = R(\va_3) = 1$
\vfill
\end{minipage}}
\\
\end{tabular}

%% file: fig2.tex
\small
\begin{tabular}{@{}l@{\hspace*{.1cm}}l@{}}
\includegraphics[width=3.4cm]{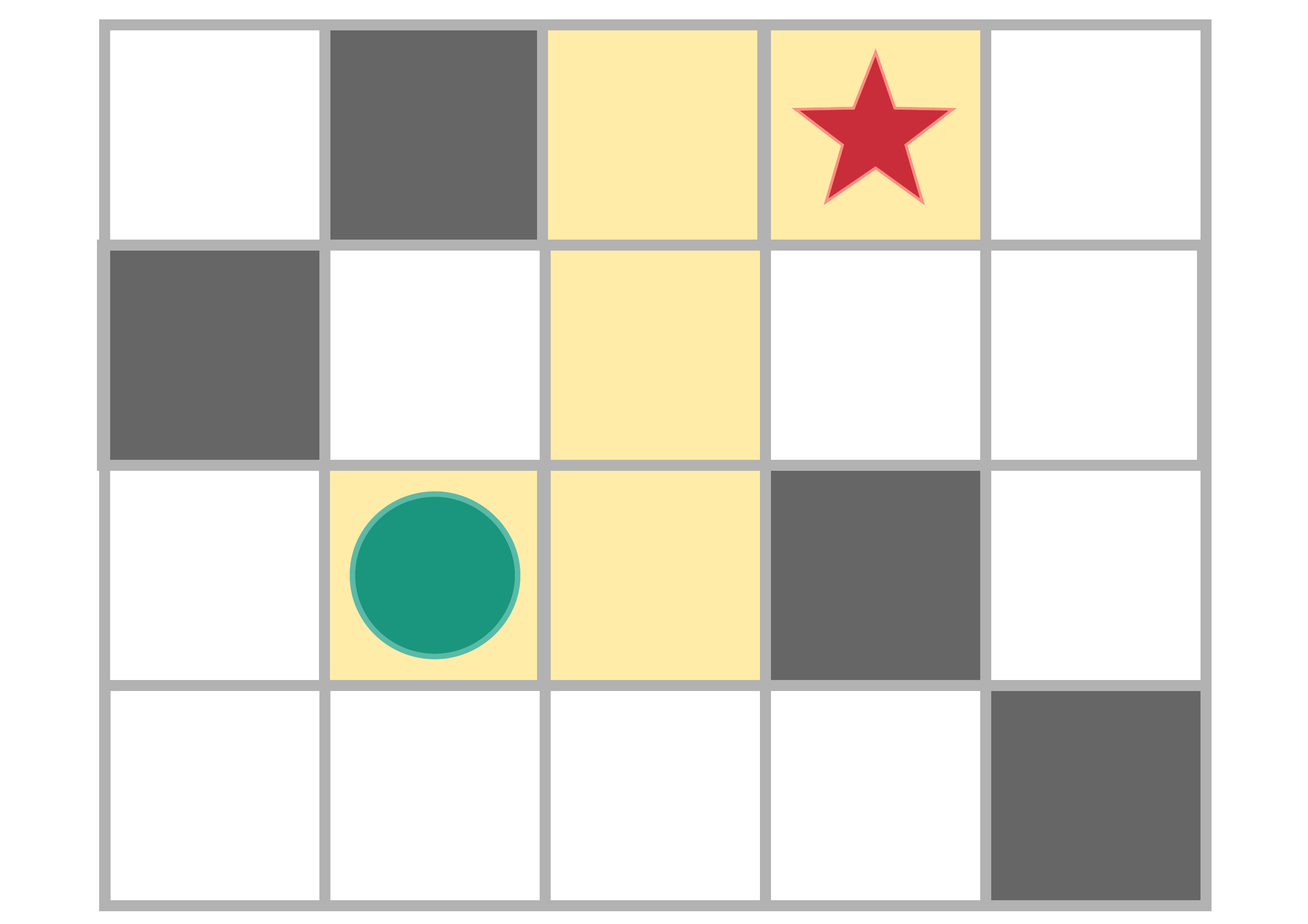}
&
\raisebox{2.15cm}{\begin{minipage}[t]{4.3cm}
$\vx$ = ``Right Up Up Right''\\
$R(\va) = \one{\texttt{Execute}(\bullet,\va) = \star}$\\[.3cm]
$\va_1 = (\rightarrow, \uparrow, \uparrow, \rightarrow)$\\
$\va_2 = (\leftarrow, \rightarrow, \rightarrow, \uparrow, \uparrow, \rightarrow)$\\
$\va_3 = (\uparrow, \rightarrow, \rightarrow, \uparrow)$\\
$R(\va_1) = R(\va_2) = R(\va_3) = 1$
\vfill
\end{minipage}}
\\
\end{tabular}

%% file: background.tex
\subsection{Problem statement}

Let $\vx$ denote a complex input, such as a natural language question
or instruction, which places an agent in some context.  Let $\va$
denote a multivariate response, such as an action trajectory that the
agent should produce. Let $R(\va \mid \vx, y) \in \{0, 1\}$ denote a
contextual success-failure feedback that uses some side information
$y$ to decide whether $\va$ is successful in the context of $\vx$ and
$y$. For instance, $y$ may be some goal specification, \eg~the answer
(denotation) in \figref{fig:fig1}, or the 2D coordinates of the goal in
\figref{fig:fig2}. For simplicity of the exposition, we assume that
$R(\va \mid \vx, y)$ is deterministic, even though our results are
applicable to stochastic rewards as well.  To simplify the equations, we
drop the conditioning of the return function on $\vx$ and $y$ and
express the return function as $R(\va)$.

Our aim is to optimize the parameters of a stochastic policy
$\pi(\va\mid \vx)$ according to a training set in order to maximize the
empirical success rate of a policy on novel test contexts. For
evaluation, the agent is required to only provide a single action
trajectory $\widehat{\va}$ for each context $\vx$, which is
accomplished via greedy decoding for interactive environments, and
beam search for non-interactive environments to perform approximate
inference:
\begin{equation}
\widehat{\va} ~\approx~\argmax{\va \in \setA(\vx)}~{\pi(\va \mid \vx)}~.
\end{equation}
Let $\setA(\vx)$ denote the combinatorial set of all plausible action
trajectories for a context $\vx$, and let $\setAp(\vx)$ denote a
subset of $\setA(\vx)$ comprising successful trajectories,
\ie~$\setAp(\vx) \equiv \{\va \in \setA(\vx) \mid R(\va\mid \vx, y)
=1\}$.

\subsection{Standard Objective Functions}
To address the problem of policy learning from binary success-failure
feedback, previous work has proposed the following objective functions:\\[.3cm]
\triangle\textbf{IML (Iterative Maximum Likelihood)} estimation
\cite{liang2017nsm,pqt2018} is an iterative process for optimizing a
policy based on 
\begin{equation}
\displaystyle O_{\mathrm{IML}} = \sum_{\vx \in \sD} \frac{1}{\lvert \setAp(\vx) \rvert}\sum_{\vyp \in \setAp(\vx)} \log \pi(\vyp\mid\vx)~.
\label{eq:iml}
\end{equation}
The key idea is to replace $\setAp(\vx)$ in \eqref{eq:iml} with a
buffer of successful trajectories collected so far, denoted
$\setBp(\vx)$. While the policy is being optimized based on
\eqref{eq:iml}, one can also perform exploration by drawing {\em i.i.d.}
samples from $\pi(\cdot \mid \vx)$ and adding such samples to
$\setBp(\vx)$ if their rewards are positive.

The more general variant of this objective function for non-binary
reward functions has been called Reward Augmented Maximum Likelihood
(RAML)~\cite{norouzi2016reward}, and one can think of an iterative
version of RAML as well,
\begin{equation}
\displaystyle O_{\mathrm{RAML}} = 
\sum_{\vx \in \sD} \frac{1}{Z(\vx)}
\!\! \sum_{\vy \in \setA(\vx)} \!\!\exp(R(\va)/\tau) \log \pi(\vy\mid\vx)~,
\label{eq:raml}
\end{equation}
where $Z(\vx) \equiv \sum_{\va \in \setA} \exp(R(\va)/\tau)$.\\[.4cm]
\triangle \textbf{MML (Maximum Marginal Likelihood)}
\cite{guu2017language,berant2013semantic} is an alternative approach to parameter
estimation related to the EM algorithm, which is only concerned with
the {\em marginal} probability of successful trajectories and not with
the way probability mass is distributed across $\setAp(\vx)$,
\begin{equation}
\displaystyle O_{\mathrm{MML}} = \sum_{\vx \in \sD} \log
\!\! \sum_{\vyp \in \setAp(\vx)} \!\! \pi(\vyp \mid \vx)~.
\label{eq:mml}
\end{equation}
Again, $\setAp(\vx)$ is approximated using $\setBp(\vx)$
iteratively. \citet{dayan1997using} also used a variant of this
objective function for Reinforcement Learning.\\[.4cm]
\triangle \textbf{RER (Regularized Expected Return)} is the common objective
function used in RL
\begin{equation}
\displaystyle O_{\mathrm{RER}} = \sum_{\vx \in \sD} \tau {\mathcal H}(\pi(\cdot \mid \vx)) +\!\! \sum_{\vy \in \setA(\vx)} R(\vy) \pi(\vy \mid \vx),
\label{eq:rer}
\end{equation}
where $\tau \ge 0$ and $\mathcal H$ denotes Shannon Entropy.  Entropy
regularization often helps with stability of policy optimization
leading to better solutions~\cite{williams1991function}.

\citet{NIPS2018_8204} make the important observation that the
expected return objective can be expressed as a sum of two terms: a
summation over the trajectories inside a context specific buffer
$\setBp(\vx)$ and a separate expectation over the trajectories outside
of the buffer:
\begin{equation}
\displaystyle O_{\mathrm{ER}} = \sum_{\vx \in \sD}
\underbrace{\sum_{\vy \in \setBp(\vx)}\!\! R(\vy) \pi(\vy \mid \vx)}_{\text{enumeration inside buffer}} + 
\underbrace{\sum_{\vy \not\in \setBp(\vx)}\!\! R(\vy) \pi(\vy \mid \vx)}_{\text{expectation outside buffer}}.
\label{eq:mapo}
\end{equation}
Based on this observation, they propose to use enumeration to estimate
the gradient of the first term on the RHS of \eqref{eq:mapo} and use
Monte Carlo sampling followed by rejection sampling to estimate the gradient of
the second term on the RHS of \eqref{eq:mapo} using the
REINFORCE~\cite{Williams92simplestatistical} estimator. This procedure
is called Memory Augmented Policy Optimization (MAPO) and in its ideal
form provides a low variance unbiased estimate of the gradient of
\eqref{eq:mapo} for deterministic $R(\cdot)$. Note that one can also
incorporate entropy into MAPO~\cite{NIPS2018_8204} as the
contribution of entropy can be absorbed into the reward function as
$R'(\va) = R(\va) - \tau\log\pi(\va\mid\vx)$. We make heavy use of the
MAPO estimator and build our code\footnote{Our open-source implementation can be found at
\url{https://github.com/google-research/google-research/tree/master/meta_reward_learning}.} on top of the open source code of
MAPO generously provided by the authors.

%% file: method1.tex
When it comes to using $\displaystyle O_{\mathrm{IML}}$
\eqref{eq:iml}, $\displaystyle O_{\mathrm{MML}}$ \eqref{eq:mml}, and
$\displaystyle O_{\mathrm{RER}}$~\eqref{eq:rer} for learning from
sparse feedback (\eg~program synthesis) and comparing the empirical
behavior of these different objective functions, there seems to be
some disagreement among previous work. \citet{pqt2018} suggest that
IML outperforms RER on their program synthesis problem, whereas
\citet{liang2017nsm} assert that RER significantly outperforms IML on
their weakly supervised semantic parsing problem.  Here, we present
some arguments and empirical evidence that justify the results of both
of these papers, which helps us develop a novel combination of IML and
RER that improves the results of \cite{liang2017nsm}.

Inspired by~\cite{norouzi2016reward,nachum2016improving}, we first
note that the IML objective per context $\vx$ can be expressed in
terms of a KL divergence between an optimal policy $\pi^*$ and the
parametric policy $\pi$, \ie~$\kl{\pi^*}{\pi}$, whereas the RER
objective per context $\vx$ can be expressed in terms of the same KL
divergence, but {\em reversed}, \ie~$\kl{\pi}{\pi^*}$. It is well
understood that $\kl{\pi^*}{\pi}$ promotes {\em mode covering}
behavior, whereas $\kl{\pi}{\pi^*}$ promotes mode seeking behavior. In
other words, $\kl{\pi^*}{\pi}$ encourages all of the trajectories in
$\setAp$ to have an equal probability, whereas RER, at least when
$\tau = 0$, is only concerned with the {\em marginal} probability of
successful trajectories and not with the way probability mass is
distributed across $\setAp(\vx)$ (very much like MML).
Notably, \citet{guu2017language} proposed an objective combining RER and MML
to learn a robust policy that can discount spurious trajectories.

\begin{figure}[t]
  \begin{center}
    \begin{tabular}{@{}c@{}c@{}}
      \includegraphics[width=.51\columnwidth]{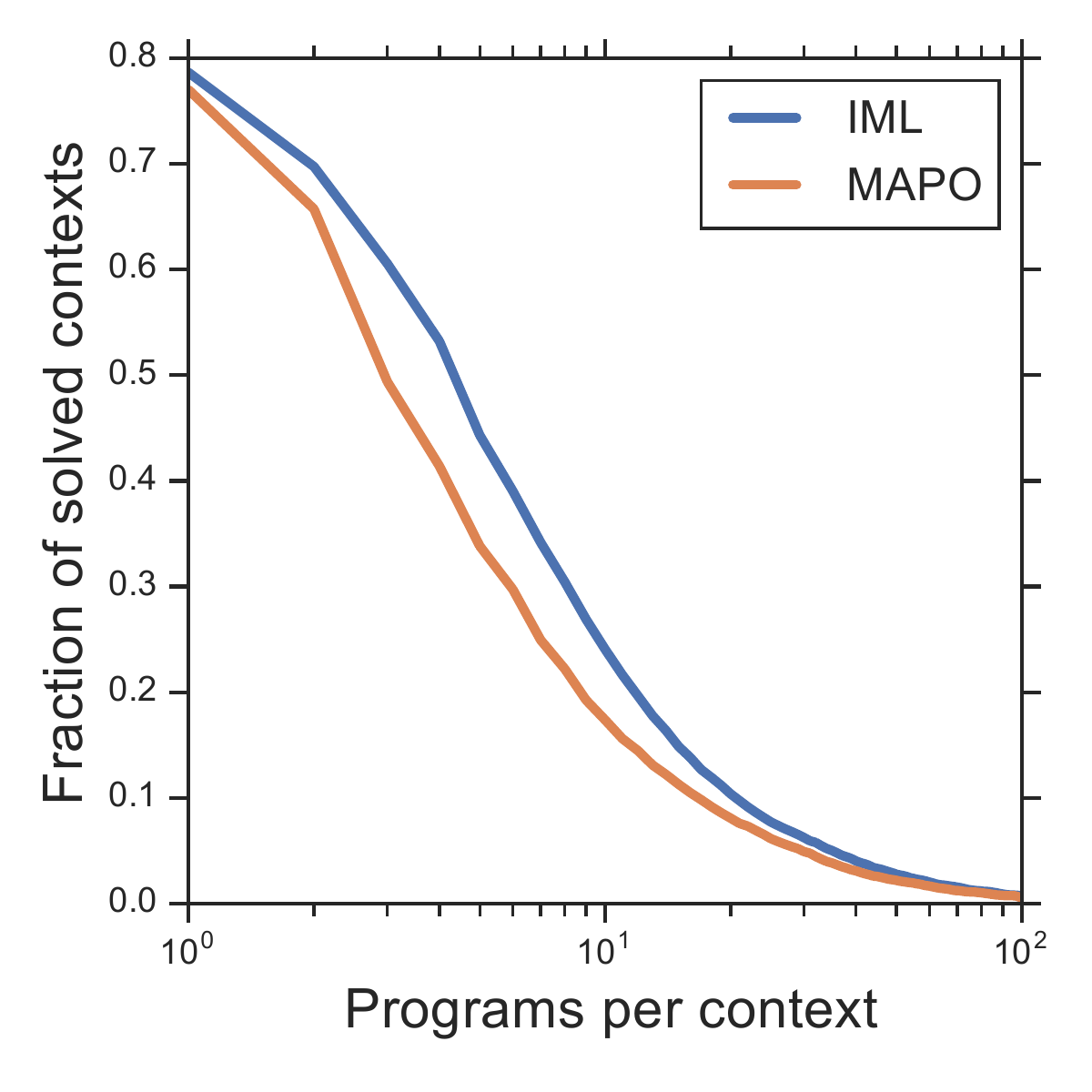} & \includegraphics[width=.51\columnwidth]{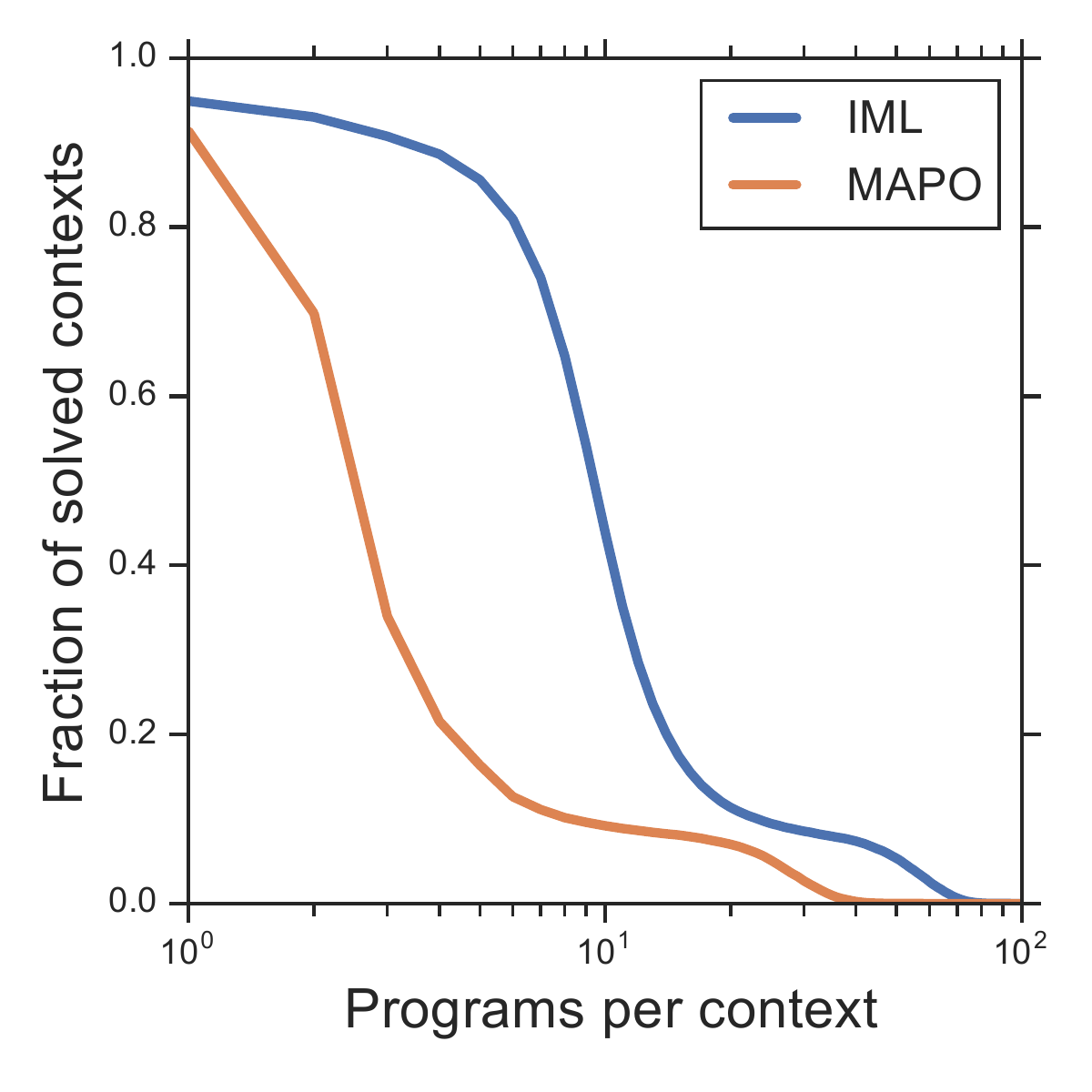} \\[-.2cm]
      (a) & (b) \\
    \end{tabular}
    \vspace*{-0.1in}
    \caption{Fraction of total contexts for which at least k programs ($\displaystyle 1 \le k \le 100$) are discovered
    during the entire course of training using the IML and MAPO (\ie~RER)
    objectives on weakly-supervised semantic parsing datasets
    (a) \textsc{WikiTableQuestions} and (b) \textsc{WikiSQL}.
    } \label{fig:mapo_iml_comparison} \end{center}
\label{fig:exploration}
\vspace*{-0.35in}
\end{figure}

Our key intuition is that for the purpose of exploration and
collecting a diverse set of successful trajectories (regardless of
whether they are spurious or not) robust behavior of RER and MML
should be disadvantageous. On the other hand, the mode covering
behavior of IML should encourage more exploratory behavior. We conduct
some experiments to evaluate this intuition, and
in \figref{fig:exploration}, we plot the fraction of contexts for
which $\lvert \setBp(\vx) \rvert \ge k$, \ie~the size of the buffer
$\setBp(\vx)$ after convergence is larger than $k$ as a function of
$k$ on two semantic parsing datasets.

Interestingly, we find that IML generally discovers many more
successful trajectories than MAPO. For example, the fraction of
context for which no plausible trajectory is found ($k=10^0$ on the
plots) is reduced by a few percent on both datasets, and for all other
values of $k > 1$, the curve corresponding to IML is above the curve
corresponding to MAPO, especially on \textsc{WikiSQL}. Examining the details of
the experiments in~\citet{pqt2018}, we realize that their program
synthesis tasks are primarily about discovering an individual program
that is consistent with a few input-output examples. In this context,
due to the presence of multiple input-output pairs, the issue of
underspecified rewards poses a less serious challenge as compared to the
issue of exploration. Hence, we believe that the success of IML in that context is
consistent with our results in \figref{fig:exploration}.

Based on these findings, we develop a novel combination of IML and
MAPO, which we call MAPOX (MAPO eXploratory). The key difference
between MAPO and MAPOX is in the way the initial memory buffer of
programs is initialized. In addition to using random search to
populate an initial buffer of programs as in~\cite{NIPS2018_8204},
we also use IML to find a large set of diverse trajectories, which are
passed to MAPO to select from. MAPOX can be interpreted as a two-stage
annealing schedule for temperature in \citet{nachum2016improving}, where one 
would use log-likelihood first ($\infty$ temperature) and then switch to expected
reward (zero temperature). In our experiments, we observe a notable gain from
this form of mode covering exploration combining the benefits of IML and MAPO.\\

%% file: method2.tex
Designing a reward function that distinguishes between optimal and
suboptimal behavior is critical for the use of RL in
real-world applications. This problem is particularly challenging when
expert demonstrations are not available. When learning from
underspecified success-failure rewards, one expects a considerable
benefit from a refined reward function that differentiates
different successful trajectories. While a policy $\pi(\va\mid\vx)$
optimized using a robust objective function such as RER and MML learns
its own internal preference between different successful trajectories,
such a preference may be overly complex. This complexity arises particularly
because the typical policies are autoregressive and only have limited
access to trajectory level features. Learning an auxiliary reward
function presents an opportunity for using trajectory level features
designed by experts to influence a preference among successful
trajectories.

For instance, consider the problem of weakly-supervised semantic
parsing, \ie~learning a mapping from natural language questions to
logical programs only based on the success-failure feedback for each
question-program pair. In this problem, distinguishing between
purposeful and accidental success without human supervision remains an
open problem. We expect that one should be able to discount a fraction
of the spurious programs by paying attention to trajectory-level
features such as the length of the program and the relationships
between the entities in the program and the question. The key
technical question is how to combine different trajectory level
features to build a useful auxiliary reward function.

For the general category of problems involving learning with underspecified
rewards, our intuition is that fitting a policy on spurious
trajectories is disadvantageous for the policy's generalization to
unseen contexts. Accordingly, we put forward the following hypothesis:
One should be able to learn an auxiliary reward function based on the
performance of the policy trained with that reward function on a
held-out validation set. In other words, we would like to learn reward
functions that help policies generalize better.  We propose two
specific approaches to implement this high level idea: (1) based on
gradient based Meta-Learning (MAML)~\cite{finn2017model}
(\algref{alg:meta_lur}) (2) using BayesOpt~\cite{snoek2012practical}
as a gradient-free black box optimizer
(\algref{alg:bayes_opt_lur}). Each one of these approaches has its own
advantages discussed below, and it was not clear to us before running
the experiments whether either of the techniques would work, and if so
which would work better.

{\bf Notation.} $\setDT$ and $\setDV$ denote the training and validation datasets
respectively. $\setBpT$ represents the training memory buffer containing successful trajectories (based on underspecified rewards)
for contexts in $\setDT$.

In this work, we employ a feature-based terminal reward function $R_{\vphi}$
parameterized by the weight vector $\vphi$. For a given context $\vx$,
the auxiliary reward is only non-zero for successful trajectories.
Specifically, for a feature vector $\vf(\va, \vx)$ for the context $\vx$ and trajectory
$\va$ and the underspecified rewards $R(\va \mid \vx, y)$:
\begin{equation}
R_{\vphi}(\va \mid \vx, y) = \vphi^{T} \vf(\va, \vx) R(\va \mid \vx, y)
.
\label{eqn:auxiliary_reward}
\end{equation}
Learning the auxiliary reward parameters determines the relative importance of 
features, which is hard to tune manually. Refer to the supplementary material for 
more details about the auxiliary reward features used in this work. 

\subsection{Meta Reward-Learning (MeRL)}
\begin{algorithm}[t]
   \caption{Meta Reward-Learning (MeRL)}
   \label{alg:meta_lur}
\begin{algorithmic}
  \STATE {\bfseries Input:} $\setDT$, $ \setDV$, $\setBpT$, $\setBpV$
  \FOR{step $t = 1, \dots, \mathrm{T}$}
   \STATE Sample a mini-batch of contexts $\setX_{\mathrm{train}}$ from $\setDT$ and $\setX_{\mathrm{val}}$ from $\setDV$
   \STATE Generate $n_{\mathrm{explore}}$ trajectories using $\pi_{\vtheta}$ for each context in $ \setX_{\mathrm{train}}$, $ \setX_{\mathrm{val}}$ and save successful trajectories to $\setBpT$, $\setBpV$ respectively
   \STATE Compute $ \vtheta^{\prime} = \vtheta \ -\ \alpha \nabla_{\vtheta} O_{\mathrm{train}}(\pi_{\vtheta}, R_{\vphi})$ using samples from ($\setBpT$, $\setX_{\mathrm{train}}$)
   \STATE Compute $ \vphi^{\prime} =  \vphi\ -\ \beta \nabla_{\vphi} O_{\mathrm{val}}(\pi_{\vtheta^{\prime}})$ using samples from ($\setBpV$, $\setX_{\mathrm{val}}$)
   \STATE Update $\vphi \leftarrow \vphi^{\prime}$, $\vtheta \leftarrow \vtheta^{\prime}$
  \ENDFOR
\end{algorithmic}
\end{algorithm}
An overview of MeRL is presented in Algorithm \ref{alg:meta_lur}. At each iteration of MeRL,
we simultaneously update the policy parameters $\vtheta$ and the auxiliary reward parameters $\vphi$.
The policy $\pi_\vtheta$ is trained to maximize the training objective $O_{\mathrm{train}}$~\eqref{eqn:merl1} computed using the training dataset
and the auxiliary rewards $R_{\vphi}$ while the auxiliary rewards are optimized to maximize the meta-training objective $O_{\mathrm{val}}$~\eqref{eqn:merl2}
on the validation dataset:
\begin{align}
  \begin{split}
  O_{\mathrm{train}}(\pi_{\vtheta}, R_{\vphi}) ={}& \sum_{\vx \in \setDT}\sum_{\vy \in \setBpT(\vx)}\!\! R_{\vphi}(\vy) \pi_{\vtheta}(\vy \mid \vx)\\
                                                  & + \sum_{\vx \in \setDT} \tau {\mathcal H}(\pi_{\vtheta}(\cdot \mid \vx))\label{eqn:merl1}
,
  \end{split}
  \raisetag{3\normalbaselineskip}\\
  O_{\mathrm{val}}(\pi) ={}& \sum_{\vx \in \setDV}\sum_{\vy \in \setBpV(\vx)}\!\! R(\vy) \pi(\vy \mid \vx)\label{eqn:merl2}
.
\end{align}

The auxiliary rewards $R_{\vphi}$ are not optimized directly to maximize the rewards on the validation set but optimized such that a policy learned by maximizing $R_\phi$ on the training set attains high underspecified rewards $R(\va \mid \vx, y)$ on the validation set. This indirect optimization is robust and less susceptible to spurious sequences on the validation set.

MeRL requires $O_{\mathrm{val}}$ to be a differentiable function of $\vphi$. To tackle this issue, we compute $O_{\mathrm{val}}$ using only samples from the buffer $\setBp_{\mathrm{val}}$ containing successful trajectories for contexts in $\sD_{\mathrm{val}}$. Since we don't have access to ground-truth programs, we use beam search in non-interactive environments and greedy decoding in interactive environments to generate successful trajectories using policies trained with the underspecified rewards. Note that $\setBp_{\mathrm{val}}$ is also updated during training by collecting new successful samples from the current policy at each step.

The validation objective is computed using the policy obtained after one gradient step update on the training objective and therefore, the auxiliary rewards affect the validation objective via the updated policy parameters $\vtheta^{\prime}$ as shown in equations \eqref{eqn:merl3} and \eqref{eqn:merl4}:
\begin{align}
  \vtheta^{\prime}(\vphi) ={}& \vtheta \ -\ \alpha \nabla_{\vtheta} O_{\mathrm{train}}(\pi_{\vtheta}, R_{\vphi})\label{eqn:merl3},\\
  \nabla_{\vphi} O_{\mathrm{val}}(\pi_{\vthetaP}) ={}& \nabla_{\vthetaP} O_{\mathrm{val}}(\pi_{\vthetaP}) \nabla_{\vphi} \vtheta^{\prime}(\vphi)\label{eqn:merl4}
.
\end{align}

\subsection{Bayesian Optimization Reward-Learning (BoRL)}
\begin{algorithm}[t]
   \caption{Bayesian Optimization Reward-Learning (BoRL)}
   \label{alg:bayes_opt_lur}
\begin{algorithmic}
  \STATE {\bfseries Input:} $\setDT$, $\setDV$, $\setBpT$
  \FOR{trial $k = 1, \dots, \mathrm{K}$}
    \STATE Sample a parameter vector $\vphi_{k}$ for $R_{\vphi_{k}}$ by optimizing the acquisition function $a_{\mathrm{M}}$ over Bayesian model M i.e. $\vphi_{k} \leftarrow \argmax{\vphi}\ a_{M}(\vphi \ |\ \setV_{1:k-1} )$
    \STATE Create a memory buffer $\setBp_{k}$ containing only the highest ranked trajectories in $\setBpT$ based on $R_{\vphi_{k}}$
    \FOR{step $t = 1, \dots, \mathrm{T}$}
      \STATE Sample batch of contexts $\setX_{\mathrm{train}}$ from $\setDT$
      \FOR{context $\vc$ in $\setX_{\mathrm{train}}$}
        \STATE Generate $n_{\mathrm{explore}}$ trajectories $\setS_{\vc}$ using $\pi_{\vtheta}$
        \STATE Save successful trajectories in $\setS_{\vc}$ ranked higher than any trajectory in $\setBp_{k}(\vc)$ based on $R_{\vphi_{k}}$
      \ENDFOR
      \STATE Update $\vtheta \leftarrow \vtheta - \alpha \nabla_{\vtheta} O_{\mathrm{train}}(\pi_{\vtheta})$ using samples from ($\setBp_{k}$, $\setX_{\mathrm{train}}$)
    \ENDFOR
    \STATE Evaluate $v_{k}$, the accuracy of policy $\pi$ on $\setDV$
    \STATE Augment $\setV_{1:k} = \{\setV_{1:k-1}, (\vphi_{k}, v_{k})\}$ and update the model M
  \ENDFOR
\end{algorithmic}
\end{algorithm}

An overview of BoRL is presented in Algorithm \ref{alg:bayes_opt_lur}. At each trial in BoRL, we sample auxiliary reward parameters by maximizing the acquisition function computed using the posterior distribution over the validation objective. After sampling the reward parameters, we optimize the $O_{\mathrm{RER}}$ objective on the training dataset for a fixed number of iterations. Once the training is finished, we evaluate the policy on the validation dataset, which is used to update the posterior distribution. BoRL is closely related to the previous work on learning metric-optimized example weights~\cite{zhao2018metric} for supervised learning.

BoRL does not require the validation objective $O_{\mathrm{val}}$ to be differentiable with respect to the auxiliary reward parameters, therefore we can directly optimize the evaluation metric we care about. For example, in non-interactive environments, the reward parameters are optimized using the beam search accuracy on the validation set $\displaystyle \sD_{\mathrm{val}}$. In this work, we use Batched Gaussian Process Bandits~\cite{desautels2014parallelizing} employing a Mat\'ern kernel with automatic relevance determination~\cite{rasmussen2004gaussian} and the expected improvement acquisition function~\cite{movckus1975bayesian}.

\subsection{MeRL \textit{vs.} BoRL}
BoRL offers more flexibility than MeRL since we can optimize any
non-differentiable objective on the validation set using BoRL but MeRL can only be used for
differentiable objectives. Another advantage of BoRL over MeRL is that
it performs global optimization over the reward parameters as compared
to the local gradient based optimization in MeRL. Notably, the
modular nature of Bayesian optimization and the
widespread availability of open source libraries for black box
optimization makes BoRL easier to implement than
MeRL. However, MeRL is much more computationally efficient that BoRL
due to having access to the gradients of the objective to
optimize. Additionally, MeRL has the ability to adapt the auxiliary
rewards throughout the course of policy optimization while BoRL can
only express reward functions that remain fixed during policy
optimization.

%% file: related.tex
The problem we study in this work as well as the proposed approach intersect with
many subfields of machine learning and natural language processing
discussed separately below.

{\bf Reward learning.}
Reinforcement learning (RL) problems are specified in terms of a
reward function over state-action pairs,
or a trajectory return function for problems with sparse feedback.
A key challenge in applying RL algorithms to real world problems
is the limited availability of a rich and reliable reward function.
Prior work has proposed to learn the reward function
(1) from expert demonstrations using inverse reinforcement
learning~\cite{abbeel2004apprenticeship,ziebart2008maximum} or
adversarial imitation learning~\cite{ho2016generative} and (2) from
human feedback~\cite{christiano2017deep,leike2018scalable,ibarz2018reward}.
Recently, these ideas have been applied to the automatic discovery of goal
specifications~\cite{xie2018few,bahdanau19}, text generation tasks~\cite{wang2018no,wu2017sequence,bosselut2018discourse} and the
optimization of reward functions~(\eg~\citet{gleave2018multi,fu19, shi2018towards})
via inverse RL.
By contrast, we aim to learn a reward function through
meta-learning to enhance underspecified rewards without using any form
of trajectory or goal demonstrations.
Another relevant work is LIRPG~\cite{zheng2018learning}, which
learns a parametric intrinsic reward function that can be added to the
extrinsic reward to improve the performance of policy gradient methods.
While the intrinsic reward function in LIRPG is trained to optimize the extrinsic reward, our reward function is trained to optimize the validation set performance through meta-learning, because our main concern is generalization.

{\bf Meta-learning.}
Meta-learning aims to design learning algorithms that can quickly
adapt to new tasks or acquire new skills,
which has shown recent success in
RL~\cite{finn2017model,duan2016rl,wang2016learning,nichol2018reptile}.
There has been a recent surge of interest in the field of meta-reinforcement
learning with previous work tackling problems such as automatically acquiring intrinsic
motivation~\cite{zheng2018learning}, discovering exploration
strategies~\cite{gupta2018meta, xu2018learning}, and adapting the nature of returns in
RL~\cite{xu2018meta}. It has also been applied to few-shot inverse reinforcement learning~\cite{xu2018few},
online learning for continual adaptation~\cite{nagabandi2018deep},
and semantic parsing by treating each query as a separate task~\cite{huang2018NaturalLT}.
Concurrent work~\cite{zou2019reward} also dealt with the problem of learning shaped
rewards via meta-learning.
Recent work has also applied meta-learning to reweight learning examples~\cite{ren2018learning}
to enable robust supervised learning with noisy labels, learning dynamic loss functions~\cite{wu2018learning}
and predicting auxiliary labels~\cite{liu2019self}
for improving generalization performance in supervised learning.
In a similar spirit, we use meta optimization to learn a reward function by maximizing the
generalization accuracy of the agent's policy. Our hypothesis is that
the learned reward function will weight correct trajectories more than
the spurious ones leading to improved generalization.

{\bf Semantic parsing.}  Semantic parsing has been a long-standing
goal for language
understanding~\cite{winograd1972understanding,zelle96geoquery,chen2011learning}.
Recently, weakly supervised semantic
parsing \cite{berant2013semantic,artzi2013weakly} has been proposed to
alleviate the burden of providing gold programs or logical forms as
annotations.  However, learning from weak supervision raises two main
challenges~\cite{berant2013semantic,pasupat2016denotations,guu2017language}:
(1) how to explore an exponentially large search space to find gold
programs; (2) how to learn robustly given spurious programs that
accidentally obtain the right answer for the wrong reason.  Previous
work~\cite{pasupat2016inferring,mudrakarta2018training,krishnamurthy2017neural}
has shown that efficient exploration of the search space and pruning
the spurious programs by collecting more human annotations has a
significant impact on final performance.  Some recent
work~\cite{berant2019explaining,cho2018adversarial} augments weak
supervision with other forms supervisions, such as user feedback or
intermediate results.  Recent RL
approaches~\cite{liang2017nsm,NIPS2018_8204} rely on maximizing
expected reward with a memory buffer and performing systematic search
space exploration to address the two challenges.  This paper takes
such an approach a step further, by learning a reward function that
can differentiate between spurious and correct programs, in addition
to improving the exploration behavior.

{\bf Language grounding.}
Language grounding is another important testbed for language understanding.
Recent efforts includes visual question answering~\cite{antol2015vqa} and
instruction following in simulated environments~\cite{hermann17,babyai}.
These tasks usually focus on the integration of visual and language components,
but the language inputs are usually automatically generated or simplified.
In our experiments, we go beyond simplified environments, and also demonstrate significant improvements
in real world semantic parsing benchmarks that involve complex language inputs.

%% file: experiments.tex
We evaluate our approach on two weakly-supervised semantic parsing
benchmarks, \textsc{WikiTableQuestions}~\cite{pasupat2015tables}
and \textsc{WikiSQL}~\cite{zhong2017seq2sql}. Note that we only
make use of weak-supervision in \textsc{WikiSQL} and therefore, our methods
are not directly comparable to methods trained using strong supervision in the form
of (question, program) pairs on \textsc{WikiSQL}. Additionally, we demonstrate
the negative effect of under-specified rewards on the generalization ability
of an agent in the instruction following task (refer to section \ref{instruction_follow}). For all our experiments, we report
the mean accuracy and standard deviation based on $5$ runs with identical hyperparameters.

\subsection{Instruction Following Task}
\label{instruction_follow}

We experiment with a simple instruction following environment in the
form of a simple maze of size $N \!\times\! N$ with $K$ deadly traps
distributed randomly over the maze. A goal located in one of the four
corners of the maze~(see Figure \ref{fig:fig2}). An agent is provided
with a language instruction, which outlines an optimal path that
the agent can take to reach the goal without being trapped. The agent
receives a reward of $1$ if it succeeds in reaching the goal within a
certain number of steps, otherwise $0$. To increase the difficulty of this task, 
we reverse the instruction sequence that the agent receives, 
\ie~the command ``Left Up Right" corresponds to the optimal trajectory of actions
$(\rightarrow, \uparrow, \leftarrow)$. 

We use a set of $300$ randomly generated environments with $(N, K) = (7, 14)$ 
with training and validation splits of $80\%$ and $20\%$ respectively. 
The agent is evaluated on $300$ unseen test environments from the same distribution. 
To mitigate the issues due to exploration, we train the agent using a fixed replay
buffer containing the gold trajectory for each environment.
For more details, refer to the supplementary material. We compare
the following setups for a MAPO agent trained with the same neural
architecture in \tabref{textworld_results}:\\[.25cm]
\triangle \textbf{Oracle Reward}: This agent is trained using the replay buffer containing only the gold trajectories.\\[.25cm]
\triangle \textbf{Underspecified Reward}: For each environment, we added a fixed number of additional spurious
trajectories~(trajectories which reach the goal without following the language instruction) to the oracle memory buffer.\\[.25cm]
\triangle \textbf{Underspecified + Auxiliary Reward}: In this case, we use the memory buffer with spurious trajectories similar
to the underspecified reward setup, however, we additionally learn an auxiliary reward function using
MeRL and BoRL~(see Algorithm \ref{alg:meta_lur} and \ref{alg:bayes_opt_lur} respectively).

\begin{table}[t]
\vspace*{-0.08in}
\caption{Performance of the trained MAPO agent with access to different type of rewards in the instruction following task.}
\label{textworld_results}
\vspace*{-0.1in}
\begin{center}
\begin{small}
\begin{tabular}{@{\ww}lcc@{\ww}}
\toprule
Reward structure & Dev & Test \\
\midrule
Underspecified & 73.0 \script{$\pm$ 3.4} & 69.8 \script{$\pm$ 2.5} \\
Underspecified + Auxiliary~(BoRL) & 75.3 \script{$\pm$ 1.6} & 72.3 \script{$\pm$ 2.2} \\
Underspecified + Auxiliary~(MeRL) & 83.0 \script{$\pm$ 3.6} & 74.5 \script{$\pm$ 2.5} \\
Oracle Reward & 95.7 \script{$\pm$ 1.3} & 92.6 \script{$\pm$ 1.0}\\
\bottomrule
\end{tabular}
\end{small}
\end{center}
\vspace*{-0.2in}
\end{table}

All the agents trained with different types of reward signal achieve
an accuracy of approximately $100\%$ on the training set. However, the
generalization performance of Oracle rewards $>$ Underspecified +
Auxiliary rewards $>$ Underspecified rewards. Using our
Meta Reward-Learning (MeRL) approach, we are able to bridge
the gap between Underspecified and Oracle rewards, which confirms our
hypothesis that the generalization performance of an agent can serve
as a reasonable proxy to reward learning.

\subsection{Weakly-Supervised Semantic Parsing}
\label{programsythesis}
On \textsc{WikiSQL} and \textsc{WikiTableQuestions} benchmarks, the
task is to generate an SQL-like program given a natural language
question such that when the program is executed on a relevant data
table, it produces the correct answer. We only have access to weak
supervision in the form of question-answer pairs~(see
Figure \ref{fig:fig1}). The performance of an agent trained to solve
this task is measured by the number of correctly answered questions on
a held-out test set.

\subsubsection{Comparison to state-of-the-art results} 
We compare the following variants of our technique with the current
state-of-the-art in weakly supervised semantic parsing, Memory
Augmented Policy Optimization~(\textbf{MAPO})~\cite{NIPS2018_8204}:\\[.25cm]
\triangle \textbf{MAPOX}: Combining the exploration ability of IML with generalization ability of MAPO, MAPOX runs MAPO starting from a memory buffer $\setBpT$ containing all the high reward trajectories generated during the training of IML and MAPO using underspecified rewards only.\\[.25cm]
\triangle \textbf{BoRL} (MAPOX + Bayesian Optimization Reward-Learning): As opposed to MAPOX, BoRL optimizes the MAPO objective only on the highest ranking	trajectories present in the memory buffer $\setBpT$ based on a parametric reward function learned using BayesOpt~(see Algorithm \ref{alg:bayes_opt_lur}).\\[.25cm]
\triangle \textbf{MeRL} (MAPOX + Meta Reward-Learning): Similar to BoRL, MeRL optimizes the MAPO objective with an auxiliary reward function simultaneously learned with the agent's policy using meta-learning~(see Algorithm \ref{alg:meta_lur}).

%% file: results.tex
\begin{table}[t]
\caption{Results on \textsc{WikiTableQuestions}.}
\label{wtq_results_ablation}
\begin{center}
\begin{small}
\begin{tabular}{@{\ww}l@{\www}c@{\www}c@{\www}c@{\www}c@{\ww}}
\toprule
& & & Improvement \\
Method &  Dev & Test & on MAPO \\
\midrule
MAPO & 42.2 \script{$\pm$ 0.6} & 42.9 \script{ $\pm$ 0.5} & --\\
MAPOX & 42.6 \script{$\pm$ 0.5} & 43.3 \script{ $\pm$ 0.4} & +0.4\\
BoRL & 42.9 \script{$\pm$ 0.6} & 43.8 \script{ $\pm$ 0.2} & +0.9\\
MeRL & 43.2 \script{$\pm$ 0.5} & \textbf{44.1} \script{ $\pm$ 0.2} & \textbf{+1.2}\\
\bottomrule
\end{tabular}
\end{small}
\end{center}
\vspace{-0.25in}
\end{table}

\begin{table}[t]
\caption{Results on \textsc{WikiSQL} using only weak supervision.}
\label{wikisql_results}
\begin{center}
\begin{small}
\begin{tabular}{@{\ww}l@{}c@{\www}c@{\www}c@{\ww}}
\toprule
& & & Improvement \\
Method &  Dev & Test & on MAPO \\
\midrule
MAPO & 71.8 \script{$\pm$ 0.4} & 72.4  \script{ $\pm$ 0.3} & --\\
MAPOX & 74.5 \script{$\pm$ 0.4} & 74.2 \script{ $\pm$ 0.4} & +1.8\\
BoRL & 74.6 \script{$\pm$ 0.4} & 74.2 \script{ $\pm$ 0.2} & +1.8\\
MeRL & 74.9 \script{$\pm$ 0.1} & \textbf{74.8} \script{$\pm$ 0.2} & \textbf{+2.4}\\
\midrule
MAPO (Ens. of 5)  & - & 74.2 & --\\
MeRL (Ens. of 5) & - & \textbf{76.9}  & \textbf{+2.7}\\
\midrule
\end{tabular}
\end{small}
\end{center}
\vspace{-0.25in}
\end{table}

{\bf Results.}~We present the results on weakly-supervised semantic
parsing in \tabref{wtq_results_ablation}
and \tabref{wikisql_results}. We observe that MAPOX noticeably
improves upon MAPO on both datasets by performing better exploration.
In addition, MeRL and BoRL both improve upon MAPOX
in \textsc{WikiTableQuestions} demonstrating that even when a diverse
set of candidates from IML are available, one still benefits from our
automatic reward learning framework. On \textsc{WikiSQL}, we do
not see any gain from BoRL on top of MAPOX, however, MeRL improves
upon MAPOX by 0.6\% accuracy. \tabref{wikisql_results} also shows
that even with ensembling $5$ models, MeRL significantly outperforms
MAPO.  Finally, \tabref{wikitable_results} compares our approach with
previous works on \textsc{WikiTableQuestions}. Note that the learned 
auxiliary reward function matches our intuition, e.g. it prefers 
programs with more entity matches and shorter length.

\begin{table}[t]
\caption{Comparison to previous approaches for \textsc{WikiTableQuestions}}
\label{wikitable_results}
\begin{center}
\begin{small}
\begin{tabular}{lccr}
\toprule
Method & Ensemble Size & Test \\
\midrule
\citet{pasupat2015tables}  & - & 37.1 \\
\citet{Neelakantan2016LearningAN} & 15 & 37.7 \\
\citet{haug2018NeuralMR} & 15 & 38.7 \\
\citet{zhang2017macro} & - & 43.7 \\
MAPO~\cite{NIPS2018_8204} & 10 & 46.3 \\
MeRL & 10  & \textbf{46.9} \\
\bottomrule
\end{tabular}
\end{small}
\end{center}
\vspace*{-0.25in}
\end{table}

\subsubsection{Utility of meta-optimization}
We compare MeRL's meta-optimization approach to post-hoc ``fixing" the policy obtained after training using underspecified rewards. Specifically, we learn a linear re-ranking function which is trained to maximize rewards on the validation set by rescoring the beam search samples on the set. The re-ranker is used to rescore sequences sampled from the learned policy at test time. We implemented two variants of this baseline: 1) Baseline 1 uses the same trajectory-level features as our auxiliary reward function, 2) Baseline 2 uses the policy's probability in addition to the auxiliary reward features in the ranking function. We use the policies learned using MAPOX for these baselines and evaluate them on \textsc{WikiTableQuestions}.

{\bf Results.}~Baseline 1 and 2 resulted in -3.0\% drop and +0.2\% improvement in test accuracy respectively, as opposed to +0.8\% improvement by MeRL over MAPOX. MeRL's improvement is significant as the results are averaged across 5 trials. These results demonstrate the efficacy of the end-to-end approach of MeRL as compared to the two stage approach of learning a policy followed by reranking to fix it. Additionally, the learned auxiliary rewards for MeRL only have to distinguish between spurious and non-spurious programs while the post-hoc reranker has to differentiate between correct and incorrect programs too.

%% file: conc.tex
In this paper, we identify the problem of learning from
{\em sparse} and {\em underspecified} rewards. We tackle this problem by employing
a mode covering exploration strategy and meta learning an auxiliary terminal
reward function without using any expert demonstrations.  

As future work, we'd like to extend our approach to
learn non-terminal auxiliary rewards as well as replace the linear reward model with more powerful models
such as neural networks. Another interesting direction is to improve upon the local optimization behavior
in MeRL via random restarts, annealing etc.

%% file: acknowledgment.tex
{\bf Acknowledgments} We thank Chelsea Finn, Kelvin Guu and anonymous reviewers for their review of the early draft of the paper and Christian Howard for helpful discussions.

%% file: sup.tex
\section{Semantic Parsing}
\label{sup:semantic}

Our implementation is based on the open source implementation of MAPO~\cite{NIPS2018_8204} in
Tensorflow~\cite{abadi2016tensorflow}. We use the same model
architecture as MAPO which combines a seq2seq model augmented by a
key-variable memory~\cite{liang2017nsm} with a domain specific
language interpreter. We utilized the hyperparameter tuning
service~\cite{golovin2017google} provided by Google Cloud for BoRL.

\subsection{Datasets}

{\bf \textsc{WikiTableQuestions}}~\cite{pasupat2015tables} contains tables extracted from Wikipedia and question-answer pairs about the tables. There are 2,108 tables and 18,496 question-answer pairs splitted into train/dev/test set. We follow the construction in~\cite{pasupat2015tables} for converting a table into a directed graph that can be queried, where rows and cells are converted to graph nodes while column names become labeled directed edges. For the questions, we use string match to identify phrases that appear in the table. We also identify numbers and dates using the CoreNLP annotation released with the dataset.

The task is challenging in several aspects.
First, the tables are taken from Wikipedia and cover a wide range of topics.
Second, at test time, new tables that contain  unseen column names appear.
Third, the table contents are not normalized as in knowledge-bases like Freebase, so there are noises and ambiguities in the table annotation.
Last, the semantics are more complex comparing to previous datasets like \textsc{WebQuestionsSP} \cite{yih2016webquestionssp}.
It requires multiple-step reasoning using a large set of functions, including comparisons, superlatives, aggregations, and arithmetic operations~\cite{pasupat2015tables}.

{\bf \textsc{WikiSQL}}~\cite{zhong2017seq2sql} is a recent large scale dataset on learning natural language interfaces for databases. It also uses tables extracted from Wikipedia, but is much larger and is annotated with programs (SQL). There are 24,241 tables and 80,654 question-program pairs splitted into train/dev/test set. Comparing to \textsc{WikiTableQuestions}, the semantics are simpler because SQL use fewer operators (column selection, aggregation, and conditions). We perform similar preprocessing as for \textsc{WikiTableQuestions}. We don't use the annotated programs in our experiments.

\begin{figure*}[t]
  \vspace{0.1in}
  \footnotesize
  \begin{center}
    \begin{tabular}{@{}lm{\columnwidth}@{}}
    \toprule
     Example & Comment\\
    \midrule
      \begin{tabular}{@{}m{\columnwidth}@{}}
    	Query nu-1167: \textbf{Who was the first oldest living president?} \\
    	MAPO: $v_0$ = (first all\_rows); $v_{\mathrm{ans}}$ = (hop $v_0$ r.president)\\
    	MeRL: $v_0$ = (argmin all\_rows r.became\_oldest\_living\_president-date); $v_{\mathrm{ans}}$ = (hop $v_0$ r.president)\\
      \end{tabular} & Both programs generate the correct answer despite MAPO's program being spurious since it assumes the database table to be sorted based on the \textit{became\_oldest\_living\_president-date} column.\\
    \midrule
      \begin{tabular}{@{}m{\columnwidth}@{}}
    	Query nu-346: \textbf{What tree is the most dense in India?} \\
    	MAPO: $v_0$ = (argmax all\_rows r.density); $v_{\mathrm{ans}}$ = (hop $v_0$ r.common\_name)\\
    	MeRL: $v_0$ = (filter\_str\_contain\_any all\_rows [u`india'] r.location); $v_1$ = (argmax $v_0$ r.density); $v_{\mathrm{ans}}$ = (hop $v_1$ r.common\_name)\\
      \end{tabular} & MAPO's program generates the correct answer by chance since it finds the tree with most density which also happens to be in India in this specific example.\\
	\midrule
	  \begin{tabular}{@{}m{\columnwidth}@{}}
		Query nu-2113: \textbf{How many languages has at least 20,000 speakers as of the year 2001?}\\
		MeRL: $v_0$ = (filter\_ge all\_rows [20000] r.2001\_\dots-number); $v_{\mathrm{ans}}$ = (count $v_0$)\\
		MAPO: $v_0$ = (filter\_greater all\_rows [20000] r.2001\_\dots-number); $v_{\mathrm{ans}}$ = (count $v_0$)\\
	  \end{tabular} & Since the query uses ``at least", MeRL uses the correct function token \textit{filter\_ge} (i.e $\ge$ operator) while MAPO uses \textit{filter\_greater} (i.e. $>$ operator) which accidentally gives the right answer in this case. For brevity, \textit{r.2001\_{\dots}-number} refers to
	  \textit{r.2001\_census\_1\_total\_population\_1\_004\_59\_million-number}.\\
	\bottomrule
    \end{tabular}
    \caption{Example of generated programs from models trained using MAPO and MeRL on \textsc{WikiTableQuestions}. Here, $v_{i}$ correponds to the intermediate variables computed
    by the generated program while $v_{\mathrm{ans}}$ corresponds to the variable containing the executed result of the generated program.} \label{fig:spurious_examples}
  \end{center}
\vspace{-0.2in}
\end{figure*}

\subsection{Auxiliary Reward Features}
In our semantic parsing experiments, we used the same preprocessing as implemented in MAPO.
The natural language queries are preprocessed to identify numbers and date-time entities. In addition,
phrases in the query that appear in the table entries are converted to string entities and the columns
in the table that have a phrase match are assigned a column feature weight based on the match.

We used the following features for our auxiliary reward for both \textsc{WikiTableQuestions} and \textsc{WikiSQL}:
\begin{itemize}
	\itemsep -0.05em
	\item $\vf_1$: Fraction of total entities in the program weighted by the entity length
	\item $\vf_2$, $\vf_3$, $\vf_4$: Fraction of date-time, string and number entities in the program weighted by the entity length respectively
	\item $\vf_5$: Fraction of total entities in the program
	\item $\vf_6$: Fraction of longest entities in the program
	\item $\vf_7$: Fraction of columns in the program weighted by the column weight
	\item $\vf_8$: Fraction of total columns in the program with non-zero column weight
	\item $\vf_9$: Fraction of columns used in the program with the highest column column weight
	\item $\vf_{10}$: Fractional number of expressions in the program
	\item $\vf_{11}$: Sum of entities and columns weighted by their length and column weight respectively divided by the number of expressions in the program
\end{itemize}

\subsection{Example Programs}
\figref{fig:spurious_examples} shows some natural language queries in \textsc{WikiTableQuestions} for which both the models trained using MAPO and MeRL
generated the correct answers despite generating different programs.

\subsection{Training Details}

\begin{table}[tb]
	\footnotesize
	\caption{MAPOX hyperparameters used for experiments in Table 2.}
	\label{wtq_hparam:1}
	\begin{center}
	\begin{tabular}{@{\ww}lccr@{\ww}}
	\toprule
	Hyperparameter & Value \\
	\midrule
	Entropy Regularization & 9.86 x $10^{-2}$ \\
	Learning Rate & 4 x $10^{-4}$ \\
	Dropout & 2.5 x $10^{-1}$ \\
	\bottomrule
	\end{tabular}
	\end{center}

	\caption{BoRL hyperparameters used in experiments in Table 2.}
	\label{wtq_hparam:2}
	\begin{center}
	\begin{tabular}{@{\ww}lccr@{\ww}}
	\toprule
	Hyperparameter & Value \\
	\midrule
	Entropy Regularization & 5 x $10^{-2}$ \\
	Learning Rate & 5 x $10^{-3}$ \\
	Dropout & 3 x $10^{-1}$ \\
	\bottomrule
	\end{tabular}
	\end{center}

	\caption{MeRL hyperparameters used in experiments in Table 2.}
	\label{wtq_hparam:3}
	\begin{center}
	\begin{tabular}{@{\ww}lccr@{\ww}}
	\toprule
	Hyperparameter & Value \\
	\midrule
	Entropy Regularization & 4.63 x $10^{-2}$ \\
	Learning Rate & 2.58 x $10^{-2}$ \\
	Dropout & 2.5 x $10^{-1}$ \\
	Meta-Learning Rate & 2.5 x $10^{-3}$ \\
	\bottomrule
	\end{tabular}
	\end{center}
\end{table}

\begin{table}[tb]
	\footnotesize
	\caption{MAPOX hyperparameters used for experiments in Table 3.}
	\label{wikisql_hparam:1}
	\begin{center}
	\begin{tabular}{@{\ww}lccr@{\ww}}
	\toprule
	Hyperparameter & Value \\
	\midrule
	Entropy Regularization & 5.1 x $10^{-3}$ \\
	Learning Rate & 1.1 x $10^{-3}$ \\
	\bottomrule
	\end{tabular}
	\end{center}

	\caption{BoRL hyperparameters used in experiments in Table 3.}
	\label{wikisql_hparam:2}
	\begin{center}
	\begin{tabular}{@{\ww}lccr@{\ww}}
	\toprule
	Hyperparameter & Value \\
	\midrule
	Entropy Regularization & 2 x $10^{-3}$ \\
	Learning Rate & 1 x $10^{-3}$ \\
	\bottomrule
	\end{tabular}
	\end{center}

	\caption{MeRL hyperparameters used in experiments in Table 3.}
	\label{wikisql_hparam:3}
	\begin{center}
	\begin{tabular}{@{\ww}lccr@{\ww}}
	\toprule
	Hyperparameter & Value \\
	\midrule
	Entropy Regularization & 6.9 x $10^{-3}$ \\
	Learning Rate & 1.5 x $10^{-3}$ \\
	Meta-Learning Rate & 6.4 x $10^{-4}$ \\
	\bottomrule
	\end{tabular}
	\end{center}
\end{table}

We used the optimal hyperparameter settings for training the vanilla IML and MAPO
provided in the open source implementation of MAPO.  One
major difference was that we used a single actor for our policy gradient implementation
as opposed to the distributed sampling implemented in Memory Augmented Program Synthesis.

For our \textsc{WikiTableQuestions} experiments reported in Table 2,
we initialized our policy from a pretrained MAPO
checkpoint (except for vanilla IML and MAPO) while for all
our \textsc{WikiSQL} experiments, we trained the agent's policy
starting from random initialization.

For the methods which optimize the validation accuracy using the
auxiliary reward, we trained the auxiliary reward parameters for a
fixed policy initialization and then evaluated the top $K$ hyperparameter settings 5
times (starting from random initialization for \textsc{WikiSQL} or on 5 different pretrained MAPO
checkpoints for \textsc{WikiTableQuestions}) and picked the hyperparameter setting with the best
average validation accuracy on the 5 runs to avoid the danger of
overfitting on the validation set.

We only used a single run of IML for both \textsc{WikiSQL}
and \textsc{WikiTableQuestions} for collecting the exploration
trajectories. For WikiSQL, we used greedy exploration with one
exploration sample per context during training. We run the best
hyperparameter setting for 10k epochs for both \textsc{WikiSQL}
and \textsc{WikiTableQuestions}. Similar to MAPO, the ensembling results reported in Table 4,
used 10 different training/validation splits of the \textsc{WikiTableQuestions} dataset.
This required training  different IML models on each split to collect the exploration
trajectories.

We ran BoRL for 384 trials for \textsc{WikiSQL} and 512 trials for \textsc{WikiTableQuestions} respectively.
We used random search with 30 different settings to obtain the optimal hyperparameter values
for all our experiments. The detailed hyperparameter settings for \textsc{WikiTableQuestions}
and \textsc{WikiSQL} experiments are listed in \tabref{wtq_hparam:1} to \tabref{wtq_hparam:3} and \tabref{wikisql_hparam:1} to \tabref{wikisql_hparam:3} respectively.
Note that we used a dropout value of 0.1 for all our experiments on \textsc{WikiSQL} except
MAPO which used the optimal hyperparameters reported by \citet{NIPS2018_8204}.

\section{Instruction Following Task}
\label{sup:ift}

\subsection{Auxiliary Reward Features}
In the instruction following task, the auxiliary reward function was computed using the single and pairwise comparison of
counts of symbols and actions in the language command $\displaystyle \vx$ and agent's trajectory $\displaystyle \vy$ respectively. Specifically, we created a
feature vector $\displaystyle \vf$of size 272 containing binary features of the form $\displaystyle f(a, c) = \#_{a}(\vx) == \#_{c}(\vy)$
and $\displaystyle \vf(ab, cd) = \#_{ab}(\vx) == \#_{cd}(\vy)$ where $ a, b \in$  \{Left, Right, Up, Down\} and $c, d \in$ \{0, 1, 2, 3\} and $\displaystyle \#_i(\vj)$ represents
the count of element $i$ in the vector $\displaystyle \vj$. We learn one weight parameter for each single count comparison feature. The weights for the pairwise features are
represented using the weights for single comparison features $\vw_{(ab,cd)} = \alpha * \vw_{ac} * \vw_{bd} +  \beta * \vw_{ad} * \vw_{bc}$ using the additional weights $\alpha$
and $\beta$.

The auxiliary reward is a linear function of the weight parameters~(see equation 7). However, in case of MeRL, we also used a softmax transformation of the linear auxiliary reward computed over all the possible trajectories (at most 10) for a given language instruction.

\subsection{Training Details}
We used the Adam Optimizer~\cite{kingma2014adam} for all the setups with a replay buffer memory weight clipping of 0.1 and full-batch training.
We performed hyperparameter sweeps via random search over the interval ($10^{-4}$, $10^{-2}$) for learning rate and meta-learning rate
and the interval ($10^{-4}$, $10^{-1}$) for entropy regularization. For our MeRL setup with auxiliary + underspecified rewards,
we initialize the policy network using the MAPO baseline trained with the underspecified rewards. The hyperparameter
settings are listed in \tabref{ift:5} to \tabref{ift:7}. MeRL was trained for 5000 epochs while other setups were trained
for 8000 epochs. We used 2064 trials for our BoRL setup which was approximately 20x the number of trials we used to tune hyperparameters
for other setups.

\begin{table}[tb]
\footnotesize
\caption{MAPO hyperparameters used for the setup with Oracle rewards in Table 1.}
\label{ift:5}
\begin{center}
\begin{tabular}{@{\ww}lcccr@{\ww}}
\toprule
Hyperparameter & Value \\
\midrule
Entropy Regularization & 3.39 x $10^{-2}$ \\
Learning Rate & 5.4 x $10^{-3}$ \\
\bottomrule
\end{tabular}
\end{center}

\caption{MAPO hyperparameters used for the setup with underspecified rewards in Table 1.}
\label{ift:6}
\begin{center}
\begin{tabular}{@{\ww}lcccr@{\ww}}
\toprule
Hyperparameter & Value \\
\midrule
Entropy Regularization & 1.32 x $10^{-2}$ \\
Learning Rate & 9.3 x $10^{-3}$ \\
\bottomrule
\end{tabular}
\end{center}

\caption{MeRL hyperparameters used for the setup with underspecified + auxiliary rewards in Table 1.}
\label{ift:7}
\begin{center}
\begin{tabular}{@{\ww}lcccr@{\ww}}
\toprule
Hyperparameter & Value \\
\midrule
Entropy Regularization & 2 x $10^{-4}$ \\
Learning Rate & 4.2 x $10^{-2}$ \\
Meta-Learning Rate & 1.5 x $10^{-4}$ \\
Gradient Clipping & 1 x $10^{-2}$ \\
\bottomrule
\end{tabular}
\end{center}
\end{table}